\documentclass[letterpaper, 9 pt, conference]{ieeeconf}  

\IEEEoverridecommandlockouts                              

\usepackage{graphics} 
\usepackage{epsfig} 
\usepackage{mathptmx} 
\usepackage{times} 
\usepackage{amsmath} 
\usepackage{amssymb}  
\usepackage{color}
\usepackage{url}
\usepackage[colorlinks,
            linkcolor=red,
            anchorcolor=blue,
            citecolor=green
            ]{hyperref}

\title{\LARGE \bf
Physical Priors Augmented Event-Based 3D Reconstruction
}

\author{Jiaxu Wang$^{1\diamondsuit}$ Junhao He$^{1\diamondsuit}$, Ziyi Zhang$^{1}$ and Renjing Xu$^{1\dagger}$
\thanks{$^\dagger$Corresponding authors; $^\diamondsuit$Co-first authors}
\thanks{$^{1}$Jiaxu Wang is PhD student with MICS Thrust,
        HKUST(GZ), Email: {\tt\small jwang457@connect.hkust-gz.edu.cn}}%
\thanks{$^{1}$Junhao He and $^{1}$Ziyi Zhang are Research Assistants with MICS Thrust,
        HKUST(GZ), Email: {\tt\small junhaohebright@outlook.com,\newline ziyizhang@hkust-gz.edu.cn}}%
\thanks{$^{1}$Renjing Xu is the professor with MICS Thrust,
        HKUST(GZ), Guangzhou, China,
        Email: {\tt\small renjingxu@ust.hk}}%
\\
}

\begin{document}
\bibliographystyle{plain}

\maketitle
\thispagestyle{empty}
\pagestyle{empty}

\begin{abstract}
3D neural implicit representations play a significant component in many robotic applications.
However, reconstructing neural radiance fields (NeRF) from realistic event data remains a challenge due to the sparsities and the lack of information when only event streams are available. In this paper, we utilize motion, geometry, and density priors behind event data to impose strong physical constraints to augment NeRF training. The proposed novel pipeline can directly benefit from those priors to reconstruct 3D scenes without additional inputs. Moreover, we present a novel density-guided patch-based sampling strategy for robust and efficient learning, which not only accelerates training procedures but also conduces to expressions of local geometries. More importantly, we establish the first large dataset for event-based 3D reconstruction, which contains 101 objects with various materials and geometries, along with the groundtruth of images and depth maps for all camera viewpoints, which significantly facilitates other research in the related fields. The code and dataset will be publicly available at \url{https://github.com/Mercerai/PAEv3d}.

\end{abstract}

\section{INTRODUCTION}
3D representations serve as the foundation for many robotic applications such as navigation, manipulation, and 3D understanding. However, images captured by standard cameras are hardly used to reconstruct entire 3D scenes on account of the lack of information under suboptimal illumination environments, especially under extreme lighting conditions including over- and under- exposures. On the other side, event cameras, as neuromorphic sensors, have been demonstrated to perform well in such environments due to their high dynamic ranges which is because each pixel in event cameras individually detects the changes of brightness and only outputs a sequence of asynchronous events composed of the polarity rather than the absolute intensities. 

Unfortunately, event-based 3D representation and reconstruction tasks remain challenging because event cameras only record relative brightness changes. Several approaches combine other devices like depth sensors or standard cameras with event cameras to reconstruct 3D scenes \cite{zuo2022devo,kueng2016low,vidal2018ultimate}. However, these methods sacrifice the advantages of event sensors, such as high temporal resolution. Other approaches tackle the problems by stereo visual odometry (VO) \cite{zhou2021event, hadviger2021feature, kueng2016low, rebecq2016evo,zuo2022devo} or SLAM \cite{kim2016real, rebecq2018emvs, zhou2018semi}. These methods only can reconstruct sparse 3D models such as point clouds. The sparsity limits their usage in many scenarios. Besides, another branch represents objects as rough templates initially, then updates their deformations to align with events \cite{nehvi2021differentiable,romero2022embodied}. Nevertheless, they rely on the initialization of templates and are only constrained by specific object categories. 

\begin{figure}
    \centering
    \includegraphics[width=1\linewidth]{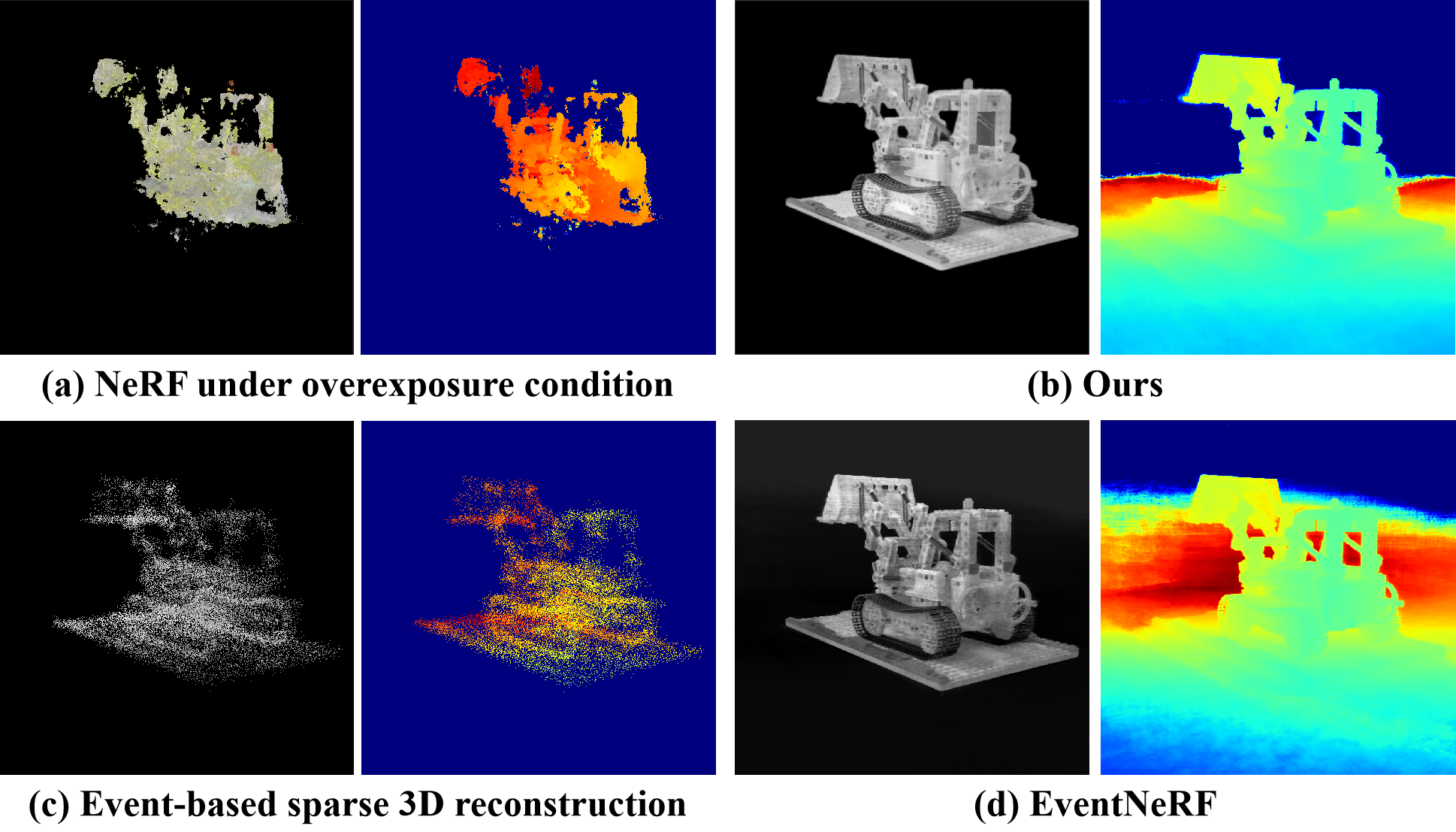}
    \vspace{-0.4cm}
    \caption{Reconstruction results of original NeRF, semi-dense point cloud from event-based approach, EventNeRF, and Ours under extreme overexposure condition. The left figure is the rendering image and the right side is the depth map.}
    \label{fig:demo}
    \vspace{-0.2cm}
\end{figure}

NeRF \cite{mildenhall_nerf_2022} gains great success in computer vision communities because it can densely represent entire 3D scenes and merely learn from images. Very recently, \cite{hwang2023ev} and \cite{rudnev_eventnerf_nodate} propose a novel paradigm that reconstructs NeRF from event streams. Even if the paradigm can perform well in certain situations, it is hard to faithfully reconstruct objects in real event data with complex geometries, textures, or realistic noises. 
Figure~\ref{fig:demo} indicates examples of reconstruction with different approaches in extreme illumination scenarios. The results are produced by overexposing the original NeRF training set and generating events with the event simulator. It can be seen the event-based approaches are less affected by the intense illumination whereas image-based NeRF is significantly destroyed. Additionally, the conventional event-based reconstruction (c) introduces severe sparsities while NeRF-based can continuously represent the 3D object. Furthermore, there is still a gap in high-quality event-based 3D reconstruction datasets. Current event-based real datasets either contain a simple handful of objects or lack corresponding image labels, which hinders the development of related tasks. 

In this paper, we tackle the problem of NeRF reconstruction from raw event data in more realistic scenes. We analyze that event data actually contain rich priors including density, motion, and geometry because events are triggered by relative movement and edges. Our aim is to embed these priors into the NeRF pipeline to physically enhance its training and improve the reconstruction quality in the aspect of textures and geometries. Moreover, we fill the gap in the lack of high-quality event-based 3D reconstruction datasets. We experimentally prove that our method outperforms the recent benchmark by a considerable margin for both synthesis and realistic datasets. Our primary technical contributions are summarized in the following:
\begin{itemize}
\item We analyze underlying motion, density, and geometry priors behind events, which we incorporate into the NeRF pipeline through the warp field, deterministic event generation model, and disparity-flow relation. 
\item We propose the probabilistic patch sampling strategy based on the spatial event density to address the local minimum optimization caused by the event sparsity, which also benefits local feature representations.  

\item We first propose a large and real event-based 3D reconstruction dataset with accurate groundtruth of image frames, foreground masks, and depth maps. The dataset contains more than 100 different objects with a wide range of materials and geometries, and would be publicly available for the community.  

\end{itemize}


\section{Related Work}
\subsection{Event-based 3D Reconstruction}
Approaches for reconstructing 3D scenes via event cameras have grown progressively, which can be roughly divided into different categories. The first branch usually uses a mix of input data modalities for additional information. Devo \cite{zuo2022devo} reconstructs 3D models with the assistance of depth sensors. \cite{levoy2023light}, \cite{cui2022dense}, and \cite{wang2022evac3d} combine intensity frames, point clouds, or IMUs with event cameras respectively. Obviously, these methods require additional sensors and cannot fully utilize all the advantages of event cameras. Second, the task can also be addressed with SLAM or VO techniques. \cite{kogler2011event, rogister2011asynchronous, camunas2014use} use a pair of synchronized event cameras to obtain the sparse 3D points. \cite{zhou2021event} solves this problem by using spatial-temporal consistency principles. Moreover, \cite{kim2016real, rebecq2018emvs, gallego2018unifying} can produce semi-dense 3D reconstructions by utilizing the knowledge prior to camera motion to integrate events over a large time interval. However, the reconstruction results of these methods are not dense enough and only contain boundaries and edges resulting in events. 
The third branch of works tends to initialize objects with 3D scans or templates, then track the deformation of the initialization to align with input events \cite{xu2020eventcap, pan2019bringing}. However, they are constrained to specific categories such as human bodies and hands. 

Inspired by the recently popular 3D dense representation NeRF, \cite{hwang2023ev} and \cite{rudnev_eventnerf_nodate} modify the traditional NeRF pipeline and make it trainable with only raw event data. These works preliminarily bridge event cameras and the neural implicit representation,  
achieving dense scene reconstruction. However, the event-based NeRF paradigm only supports recovering objects with simple geometries and regular textures because event data only contain relative illumination changes and much less information compared to the standard image, which causes ambiguity in the solution subspace. To reduce ambiguity, we propose the physically augmented event-based NeRF via underlying priors behind raw event data. Experiments illustrate that our proposed prior-augmented NeRF paradigm outperforms the benchmark considerably, especially in realistic scenarios.

\subsection{3D Scene Representations} 
Previous works have explored many different representations for modeling 3D scenes in various vision, graphics, and robotic applications. Traditional methods based on explicit representations such as point cloud \cite{qiPointnetDeepLearning2017, achlioptasLearningRepresentationsGenerative2018, liuNeuralRenderingReenactment2019}, mesh \cite{wangPixel2meshGenerating3d2018, thiesDeferredNeuralRendering2019, liuGeneralDifferentiableMesh2020}, and voxel \cite{lombardiNeuralVolumesLearning2019, sitzmannDeepvoxelsLearningPersistent2019} have inherent limitations of fixed topological structures and poor quality of novel view synthesis. To address these, 3D implicit scene representations \cite{thies_deferred_2019, zhou_stereo_2018, lombardi_neural_2019, mildenhall_nerf_2022,liu_learning_2019,niemeyer_differentiable_2020,liu_dist_2020} have been presented. For instance,
DIST \cite{liu_dist_2020} and DVR \cite{niemeyer_differentiable_2020} propose differentiable rendering formulations for implicit representations. But they require explicit extraction of surface information. 

Recently, the Neural Radiance Field \cite{mildenhall_nerf_2022} has gained many successes. NeRF encodes a continuous volume representation of shape and color in the weights of an MLP, it supports efficient learning at arbitrary resolutions and enables rendering novel views with high-fidelity detail.
The superiorities of NeRF inspire subsequent works in a wide variety of robotic applications, such as robotic policies \cite{zhou_nerf_nodate, lee_uncertainty_2022,dai2023graspnerf,byravan2023nerf2real,lin2023parallel,sunderhauf2023density}, SLAM \cite{wang_co-slam_nodate, zhu_nicer-slam_2023, chung2023orbeez}, large scene reconstruction \cite{mi_switch-nerf_2023,xiangli_bungeenerf_2023,tancik_block-nerf_2022}, robotic localization \cite{zhu2023latitude,maggio2023loc,liu2023nerf}, and robotic safety \cite{tong2023enforcing}. 
SPARTN \cite{zhou_nerf_nodate}augments robot trajectories via NeRF to improve the robustness of the grasping strategy. For street view or aerial images, Block-NeRF \cite{tancik_block-nerf_2022}, Switch-NeRF \cite{mi_switch-nerf_2023}, and BungeeNeRF \cite{xiangli_bungeenerf_2023} enable city level reconstruction. NICER-SLAM \cite{zhu_nicer-slam_2023} uses monocular geometric cues and optical flow as supervisions to optimize hierarchical neural implicit representation for building the SLAM system.

\begin{figure*}[t]
    \centering
    \includegraphics[width=1\linewidth]{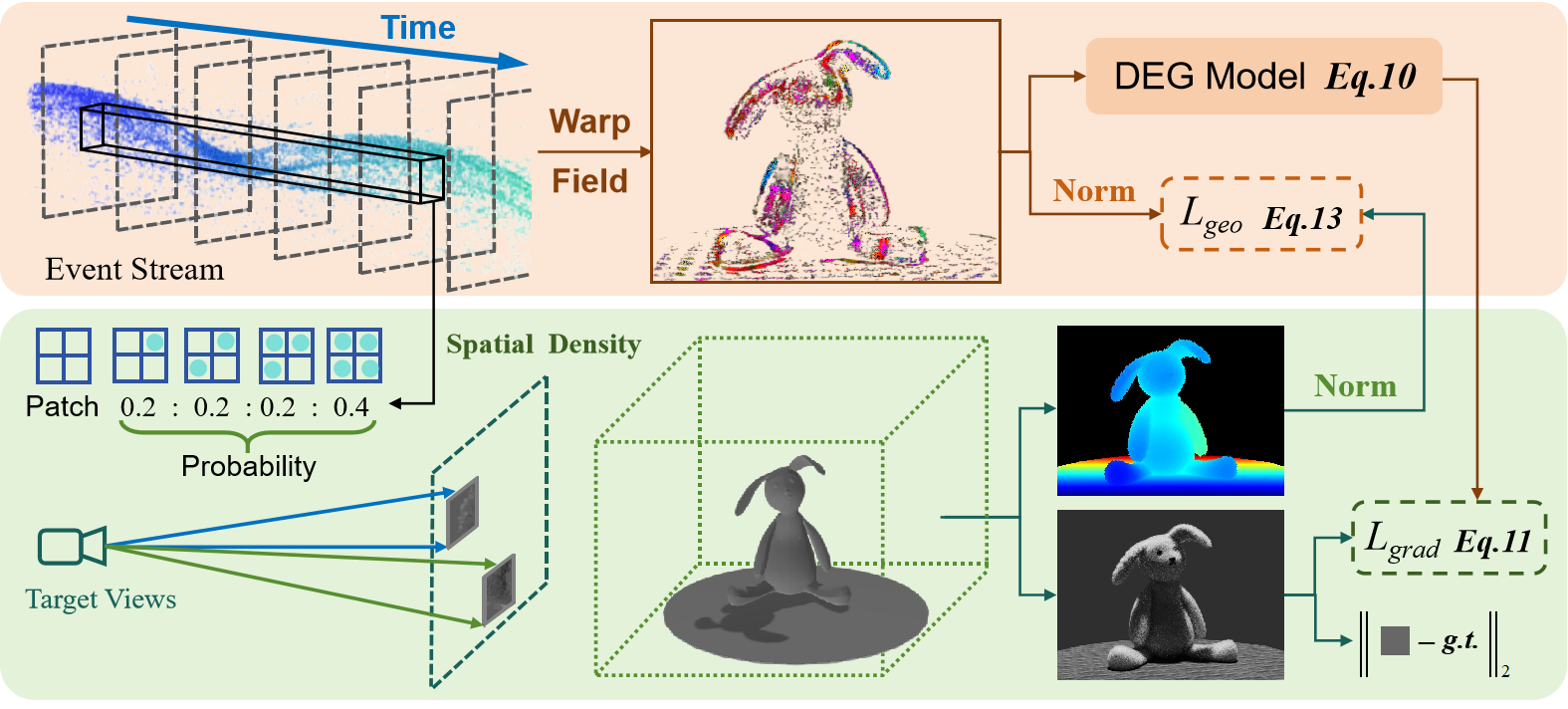}
    \vspace{-0.6cm}
    \caption{The whole pipeline of the proposed approach. There are two main branches, i.e. the prior extraction and NeRF rendering branches. The priors are incorporated into the NeRF pipeline at sampling and loss parts. }
    \label{fig:pipeline}
    \vspace{-0.6cm} 
\end{figure*}

\section{Method}
First, we provide an overview diagram of our full method in Figure~\ref{fig:pipeline} which contains all main components. 
\subsection{Preliminary Background}
This section introduces and analyzes the preliminary knowledge about volumetric rendering and event-based radiance fields. Conventional radiance fields store scene information in the parameters of MLPs and one can explore the scene by repeatedly accessing the neural network that is defined by $c,\sigma=f(x,y,z,d)$. The color $c$ is not only dependent on locations but also the direction from which we observe. The opacity can be interpreted as the occupancy rate of this location. After these definitions, volumetric rendering can be applied to composite the results of sampled points on a ray into a pixel, as in Equation~\ref{eq.vol_rendering}.  

\begin{equation}
    C(r) = \sum_{t=t_n}^{t_f}W(\sigma(t) ,t)c(t,d)
\label{eq.vol_rendering}
\end{equation}

\noindent where $r$ refers to a sampled ray, $c$ is the color for each point on the ray. $W$ is the blending weight calculated from the opacity by Equation~\ref{eq.blend weight}. 
\vspace{-0.2cm} 
\begin{equation}
    W(\sigma ,t)=\sum_{t=t_0}^{T}[(1-exp(\sigma (t)\delta))T(t)]
\label{eq.blend weight}
\end{equation}

\noindent in which $T(t)=exp(-\textstyle \sum_{s=t_0}^{t}\sigma(s)\delta))$, $\delta$ represents the distance between adjacent points on the same ray. Besides, we can approximate the depth value at a certain pixel by the following equation:
\vspace{-0.2cm} 
\begin{equation}
    D(r) = \sum_{t=t_0}^{T}W(\sigma(t) ,t)z(t)
\label{eq.depth rend}
\end{equation}

\noindent However, such NeRF model is hardly trained with pure event streams directly \cite{rebecq2019high}.  

A simple self-supervised paradigm for NeRF reconstruction from a single event stream is first proposed by \cite{hwang2023ev} and \cite{rudnev_eventnerf_nodate}. It considers accumulated event frames as the intensity changes between two images. The event-based paradigm computes losses between a pair of images as in the following:
\vspace{-0.1cm} 
\begin{equation}
    \triangle \hat L(u) = log(\hat I_{t}+b)-log(\hat I_{t-1}+b)
\label{eq.render interval}
\end{equation}

\noindent in which $I_t$ denotes the rendering image at time t, b is an infinitesimal number to prevent $log(0)$. The difference is defined in the log domain due to the principle of event cameras. The event frame is accumulated as follows: 
\vspace{-0.1cm} 
\begin{equation}
    \Delta L(u)=\sum_{t\in\triangle t}p_kC\delta (u-u_k), u\in (X,Y) 
\label{eq.event frame}
\vspace{-0.1cm} 
\end{equation}

In this equation, $p_k$ denotes event polarities, C represents the event number, and $\delta$ is the impulse function. $u$ stands for pixel coordinates. It sums all events within the time interval on the frame. The training loss function is defined as:
\vspace{-0.15cm} 
\begin{equation}
    L_{event}=||\Delta \hat L-\Delta L||^2_2 
\label{eq.loss func}
\vspace{-0.1cm}  
\end{equation}

However, this paradigm cannot recover correct 3D scenes with complex geometries, irregular textures, or realistic noises. We incorporate strong priors into the training pipeline to improve the reconstruction quality of realistic data. This will be introduced in the next section. 

\subsection{Extraction of motion priors}
Events are triggered by apparent motions parallel to the intensity gradient because event cameras respond to the apparent motion of edges. Since the event stream only reports pixel-wise brightness changes, it carries rich information about motion priors that contain relative motions, underlying geometric structures, and appearance gradients for events to be generated. These implicit priors are beneficial for 3D reconstruction from pure event streams and improve the details of geometries. 

The problems of extraction of these priors can be converted to establish data association among events. Data association is to determine which events were triggered by the same edge. All events can be warped back along their trajectories into a reference view with a timestamp $t_{ref}$ to obtain a sharper edge image. Therefore, the data association implicitly contains priors like optical flows, motion patterns, and depths. Next, we introduce how to extract those priors and how to use them to enhance the quality of reconstruction. 

Since we only focus on the reconstruction of static scenes, motion information where the effect of camera motion can be described by a homography $\theta$. Assume we are given a set of events $\varepsilon=\left \{ e_i^N \right \}$, a general function for warping events is defined as $x_k^{'}=W(x_k,t_k;\theta )$, which warps event $e_k=(x_k,t_k,p_k)$ to $e_k=(x_k^{'},t_{ref},p_k)$ according to the motion parameter $\theta$. Then following Equation~\ref{eq.event frame}, an image patch of warped events is built. The resulting sum is denoted by $H(x)$, which measures how well the events agree with the candidate trajectories. After that, we compute the variance of $H$ by the following equation:
\vspace{-0.2cm}
\begin{equation}
    V_H=\frac{1}{N_e}\sum_{i,j}(h_{ij}-u_H)  
\label{eq. variance}
\vspace{-0.2cm}
\end{equation}
\noindent where $N_e$ is the total number of pixels of H, $u_H$ is the mean of H. It is clear that correcter motion parameters result in sharper event accumulated maps, i.e. the maximum variance. Thus we optimize $\theta = argmax_\theta V_H$ using gradient ascending to solve for the warp field. One by-product is that a compensated edge map can also be obtained from this optimization.

\subsection{Learning from motion priors}
We incorporate the prior warp field, which essentially encodes the per-event optical flow, event density, and a deterministic generative event model into the NeRF training. First, the process of event generation within a given time interval $\triangle t$ can be described as:
\begin{equation}
    |\Delta L(u)|=|L(u,t)-L(u,t-\Delta t)|\ge C
\label{eq.H variance}
\end{equation}
\noindent in which C is a preset threshold. If we substitute the brightness consistency assumption (Equation~\ref{eq:bright_conti}) into its Taylor's approximation, we can get \ref{eq:event generation model}
\begin{equation}
    \frac{\delta L}{\delta t}(\textbf{u},t) + \nabla L(\textbf{u},t) \cdot \textbf{v}(\textbf{u})) = 0 
    \label{eq:bright_conti}
    \vspace{-0.1cm}
\end{equation}

\begin{equation}
   \Delta L(\textbf{u}) \approx \frac{\delta L}{\delta t}(\textbf{u},t)\Delta t = -\nabla L(\textbf{u}) \cdot \textbf{v}(\textbf{u})\Delta t
    \label{eq:event generation model}
    \vspace{-0.1cm}
\end{equation}

\noindent The left side of the equation can be regarded as event accumulation frames (defined by Equation~\ref{eq.event frame}) in a $\Delta t$. On the right side, $\textbf{v}(u)$ refers to the velocity on the trajectories, which can be obtained by accessing the warp field. Additionally, $\nabla L$ is the intensity gradient in the log domain. This term can be considered as the self-supervised target and obtained from the NeRF model. To leverage this supervision, we modified two details in the NeRF pipeline. First, conventional NeRF randomly selects some rays in a batch for training. Nevertheless, isolated sampling pixels are not allowed to compute gradients. Instead, we use patch-based random sampling to randomly select several patches and compute their gradients. Then we substitute gradients to Equation~\ref{eq:event generation model} to build the l1 gradient loss:

\begin{equation}
    L_{grad}=\frac{1}{N}\sum_p ||\nabla  L_p/\Delta t- \Delta L_p(u) \cdot v_p||_1
\label{eq:grad loss}
\vspace{-0.1cm}
\end{equation}
\noindent in which p refers to each independent patch.

\begin{figure*}[t]
    \centering
    \includegraphics[width=1\linewidth]{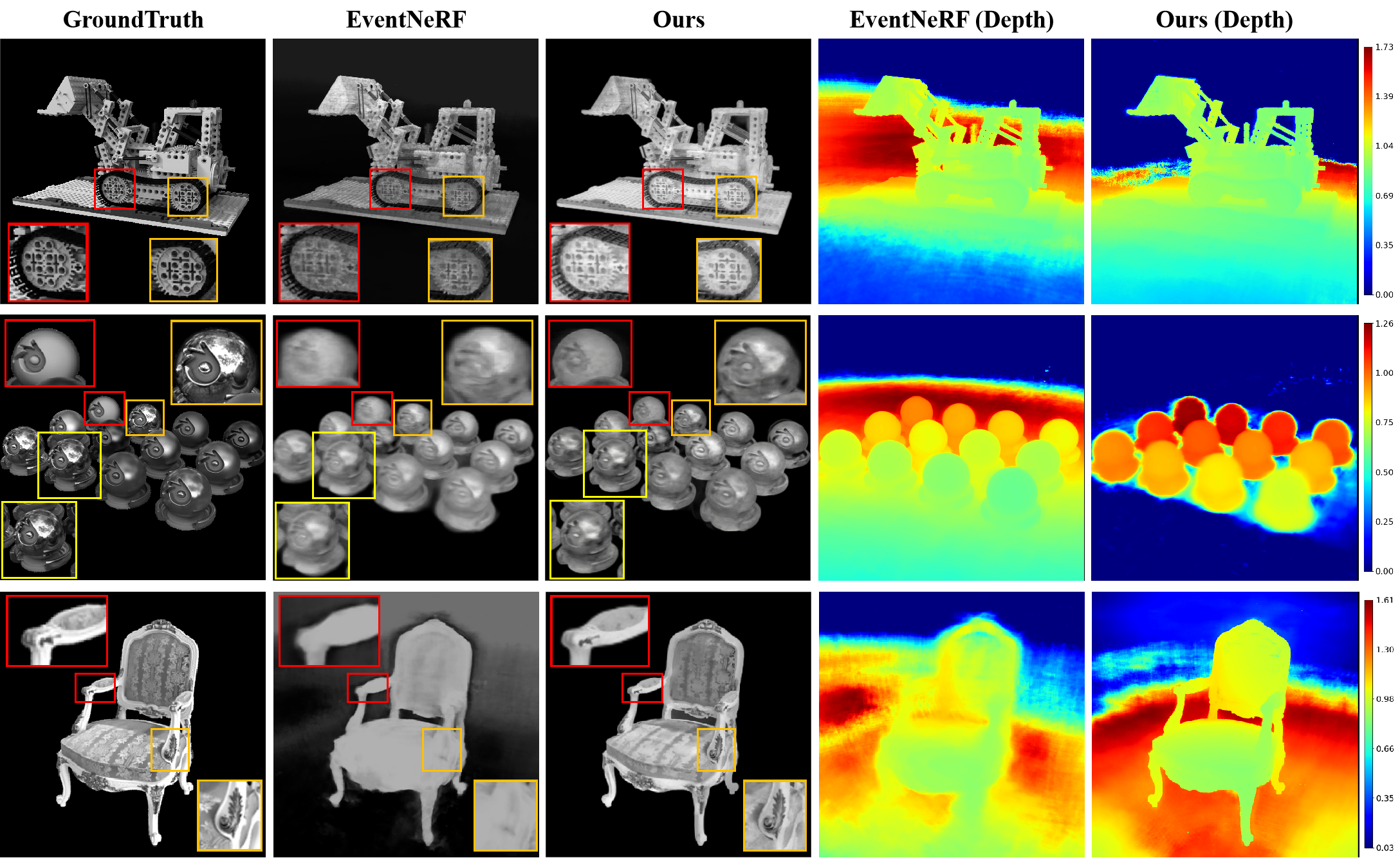}
    \vspace{-0.2cm}
    \caption{Qualitative Comparisons between ours and benchmark on the synthetic dataset.}
    \label{fig:result nerf}
    \vspace{-0.2cm}
\end{figure*}

For each independent patch, it is a square of size \(n\). We use the event spatial density to guide the sampling processes. Patches are sampled with weights of the number of pixels containing events, which we call event pixels, (the number of pixels ranges from 0 to \(n^{2}\)). However, we observed from experiments that if we evenly sample patches with different event numbers, this could lead to the exclusion of certain event patches from the sampling process. Hence we propose to sample patches with different event numbers in an uneven manner. In detail, we define a probability for each patch being sampled according to its number of pixels containing events. The probability of patches with the most event pixels reaches the maximum, whereas patches without events have the least confidence. In Equation~\ref{eq:sampling stategy}, $n_e$ denotes the number of event pixels, $f$ is a monotone increasing function and here we simply choose linear function as $f$. It is noted that other functions may reach the same effect. 

\begin{equation}
    P=\frac{1}{\sum_{0}^{n^2}f(n_e)}f(n_e) 
\label{eq:sampling stategy}
\end{equation}

This density prior not only ensures the sampling of a sufficient number of event pixels but also captures local features present within sparse event patches, thereby significantly enhancing the quality of the final texture reconstruction.

\begin{figure*}
    \centering
    \includegraphics[width=1\linewidth]{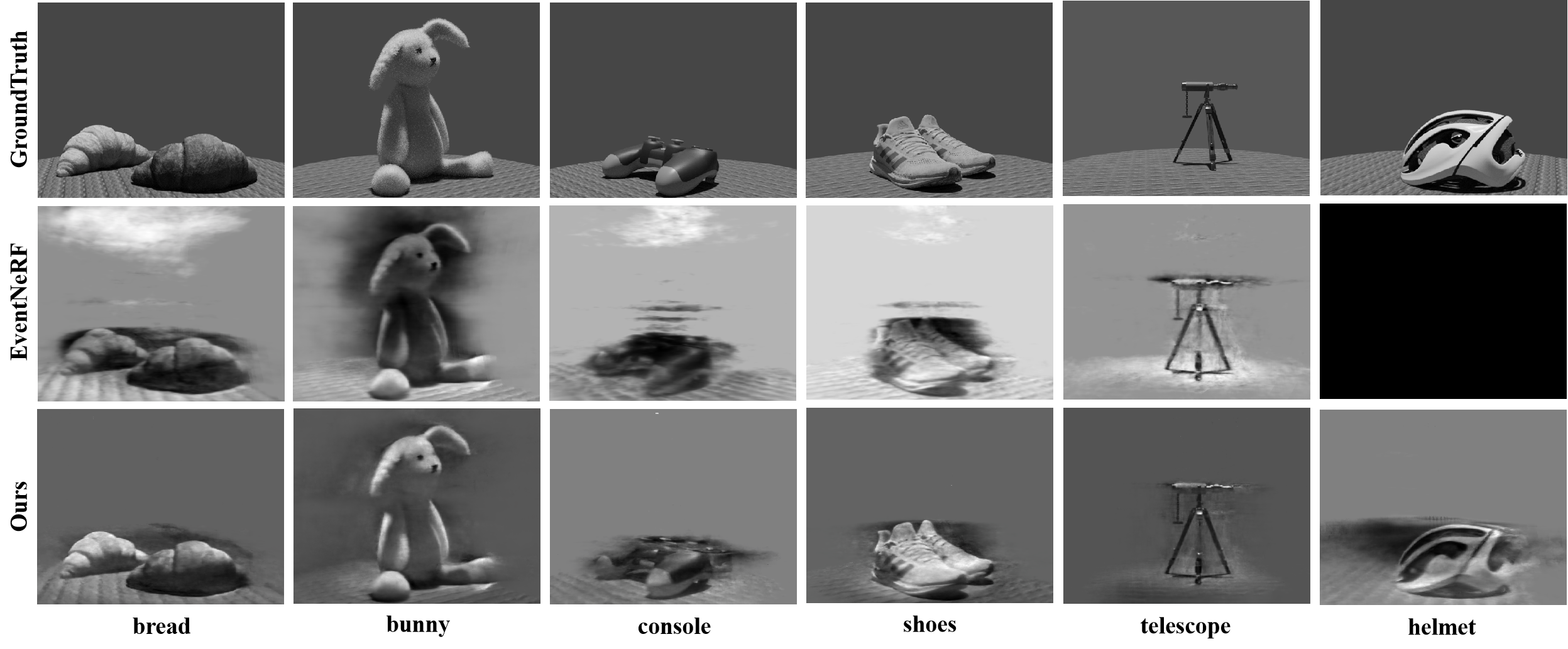}
    \vspace{-0.35cm}
    \caption{Qualitative Comparisons between ours and benchmark on the realistic dataset.}
    \label{fig:real results}
\end{figure*}

Additionally, the warp field also implies geometric priors of the scene because the scene is static and all events are caused by moving cameras. A common sense is that objects closer to cameras lead to larger displacement on the camera plane. The optical flows are roughly proportional to the disparities when only the camera is translationally moving. If we divide the camera trajectories into infinitesimal segments, each can be approximated as a translation. We use such features to regularize the geometries learned in NeRF. For each batch, we have the equation below

\begin{equation}
    L_{geo} = \frac{1}{\hat{N} } \sum_{u\in \varepsilon }^{\hat{N} } ||\frac{F_u}{F_{max}}-\frac{1}{D_uD_{min}}||_1  
\label{eq:geo loss}
\end{equation}

\noindent In this equation, $u\in \varepsilon$ represents the $(x, y)$ of sampled rays origins that contain at least one event. $F_u$ denotes the warping flow that can be attained by accessing the warp field ($F_u = W(u, t;\theta)$). $F_{max}$ is the maximum value over all $F_u$ in this batch. $D_u=D_u^{'}+\epsilon$, where $D_u^{'}$ is the output of NeRF models and can be computed via Equation~\ref{eq.depth rend}, $\epsilon$ is a very small number to prevent the divisor from reaching zero. We normalize each $F_u$ and $D_u$ over the batch to reduce the influence of noises and the infinitesimal assumption. This loss directly regularizes the learning of geometries and efficiently improves the reconstruction of details. 

%

\subsection{Implementation details}
We use the original NeRF framework containing the coarse and fine models in accordance with the benchmark. Then we compose the above loss functions to guide the model training. The total losses are described as:

\begin{equation}
    L_{total} = \alpha L_{event} + \beta L_{geo} + \gamma L_{grad}
\label{eq:total loss}
\end{equation}
\noindent where $\alpha$, $\beta$ and $\gamma$ denote the weights to balance losses. In all experiments, we uniformly choose $\alpha=1$, $\beta=0.01$, $\gamma=0.01$. For the density-based patch sampling, we fix $n$ in Equation~\ref{eq:sampling stategy} as 2. Adam optimizer with 1e-5 learning rate and $\beta_1=0.9, \beta_2=0.999$ is used to train the model. We train our model on two NVIDIA 3090 GPUs with 150,000 steps, whereas train the benchmark with 500,000 on the same devices because our model converges more quickly due to the assistance of priors. 

The depth maps rendered by NeRF usually display anomalies, and substantial inaccuracies may occur. Since the event depth is influenced by event noise, if we simply use the noisy event depth prior and the depth map rendered by NeRF for training, it can introduce significant errors, affecting the results of texture and geometric reconstruction. Therefore, we use Ambient Occlusion (AO) \cite{Tosi_2023_CVPR} to measure the confidence of the depth map rendered by NeRF.
\vspace{-0.15cm}
\begin{equation}
    AO = \sum_{i=1}^{N}T_i\alpha_i,\qquad \alpha_i = 1 - \exp(-\sigma_i\delta_i)
\end{equation}
Different from \cite{Tosi_2023_CVPR}, we start with an AO initial value of 0 and gradually increase it to a predetermined maximum value as the number of training steps increases. This is because the NeRF trained with sparse event frames has a relatively poor fit when the number of training steps is not high. 
Therefore, we employ a warm-up filtering mechanism to gradually filter out geometric loss.


\section{Experiments}

\textbf{Dataset}. We test our methods and the benchmark (EventNeRF) on both synthetic and realistic datasets. Notably, EvNeRF and EventNeRF both worked at the same time and they follow a very similar paradigm, thus we only test the EventNeRF by its official source code. For the synthetic dataset, we simply transfer the original NeRF dataset to event streams by the video2event algorithm \cite{v2e}. However, there is a large gap between synthetic and real event data in the aspect of event distributions, noises, and resolutions. Therefore, we mainly concentrate on realistic data.

Currently, there is no high-quality 3D reconstruction dataset captured by real event cameras, which obstacles the development and evaluations of algorithms, therefore we present a large dataset for event-based 3D reconstruction, which includes 101 different objects and scenarios. This dataset is recorded by a realistic DVXplore event camera and contains various materials and textures, such as plush, leather, alloy, wood, flour, metal, plastics, etc. Moreover, we provide the groundtruth of intensity frames, depth maps, and foreground masks associated with all camera poses. Although EventNeRF offers a few real event data, they only contain 10 objects and do not include the groundtruth of frames and depth maps, which cannot be used for quantitative analysis. As the space is limited, we selectively established six different scenarios including "bread", "bunny", "console", "helmet", "shoes", and "telescope". The whole dataset will be publicly available for the convenience of other researchers. 

\textbf{Metrics}. As event data reflect the relative differences in intensity rather than absolute brightness, we normalize the rendering results to align with the groundtruth before computing metrics. Then we compute PSNR, SSIM, and LPIPS for comparisons.

\begin{table}[]
\caption{Metrics of ours and the benchmark on synthetic data.}
\vspace{-0.1cm}
\renewcommand{\arraystretch}{1.3}
\setlength\tabcolsep{1.9mm}{
\begin{tabular}{ccccccc}
\hline\hline
                               & \multicolumn{3}{c}{EventNeRF} & \multicolumn{3}{c}{Ours} \\ 
& PSNR↑  & SSIM↑ & LPIPS↓ & PSNR↑ & SSIM↑ & LPIPS↓ \\ \hline
{lego}      & 21.91    & 0.921    & 0.073    & 24.51  & 0.961  & 0.067  \\
{materials} & 18.64    & 0.915    & 0.135    & 23.47  & 0.945   & 0.092  \\
{chair}     & 20.74    & 0.917    & 0.121    & 23.61  & 0.972  & 0.112  \\ \hline\hline
\end{tabular}}
\label{tab. evaluation on synthetic}
\end{table}

\begin{table}[]
\caption{Metrics of ours and the benchmark on realistic data.}
\vspace{-0.1cm}
\renewcommand{\arraystretch}{1.3}
\setlength\tabcolsep{0.9mm}{
\begin{tabular}{cccccccc}
\hline\hline
\multicolumn{1}{l}{}          &        & bread & bunny & console & shoes & telescope & helmet \\ \hline
                              & PSNR↑   & 18.59 & 22.60 & 17.35   & 20.03 & 19.63        & N/A    \\
\multicolumn{1}{l}{EventNeRF} & SSIM↑  & 0.840 & 0.968 & 0.783   & 0.870 & 0.944        & N/A    \\
\multicolumn{1}{l}{}          & LPIPS↓ & 0.318 & 0.085 & 0.304   & 0.224 & 0.280        & N/A    \\ \hline
                              & PSNR↑   & 27.15 & 25.82 & 23.27   & 22.51 & 24.74      & 22.40  \\
Ours                          & SSIM↑  & 0.962 & 0.923 & 0.950   & 0.911 & 0.935      & 0.944   \\
                              & LPIPS↓ & 0.067 & 0.017 & 0.116   & 0.059 & 0.142      & 0.063  \\ \hline\hline
\end{tabular}}
\label{tab. evaluation on real data}
\vspace{-0.3cm}
\end{table}

\subsection{Evaluation on Synthetic Event Data}
We first evaluate the proposed method on the NeRF synthetic data. We only report the comparison results on "lego", "material", and "chair" on account of the limited space. Our main focus is on realistic event data. 
It is observed that our method maintains correct structures, especially at geometry discontinuities, such as the wheel of the "lego" and the depression in the "material". The counterpart method causes severe background noises at the depth map, whereas our method attains clearer depths. Moreover, the proposed approach delivers better contrast in the renderings, while images produced by the benchmark are blurry. The quantitative and qualitative comparisons are shown in Table~\ref{tab. evaluation on synthetic} and Figure~\ref{fig:result nerf} respectively.

\subsection{Evaluation on Realistic Event Data}
The proposed dataset includes more than 100 objects and we randomly select seven items with different materials for establishment. Event NeRF only receives the 3D coordinates and view direction as input, and considers the event stream as learning targets. However, as stated in Figure~\ref{fig:pipeline}, we additionally input event streams into an optimization-based prior extraction branch to build priors that guide sampling and training procedures. This intuitively will decrease the training speed. But we precompute the warp field for each camera pose timestamp. Hence, we only compute them once during training. In "helmet", EventNeRF cannot converge properly, thereby learning nothing from event streams. Our approach faithfully reconstructs the main structures of objects even if there are some fog noises around them.  In "bread", "telescope", and "console", the benchmark infers wrong geometries and leads to large variances in depth predictions. In contrast, ours maintains relatively clean shapes and sharper boundaries. In "bunny" and "shoes", the benchmark incorrectly predicted the material to be translucent, while our method gives higher confidence to the geometry. All the above results are shown in Figure~\ref{fig:real results}. We additionally compute the metrics for the above six scenes and the results are listed in Table~\ref{tab. evaluation on real data}.

\begin{table}[]
\caption{Ablation studies for all components of losses.}
\renewcommand{\arraystretch}{1.2}
\centering
\setlength\tabcolsep{5.5mm}{
\begin{tabular}{cccc}
\hline\hline
\multicolumn{1}{l}{} & PSNR↑  & SSIM↑ & LPIPS↓ \\ \hline
$Ours_{wo/geo}$      & 20.38 & 0.865 & 0.201 \\
$Ours_{wo/grad}$     & 19.49 & 0.820 & 0.199 \\
Ours                 & 21.31 & 0.910 & 0.167 \\ \hline\hline
\end{tabular}}
\label{tab.abl1}
\end{table}

\begin{figure}[t]
    \centering
    \includegraphics[width=1\linewidth]{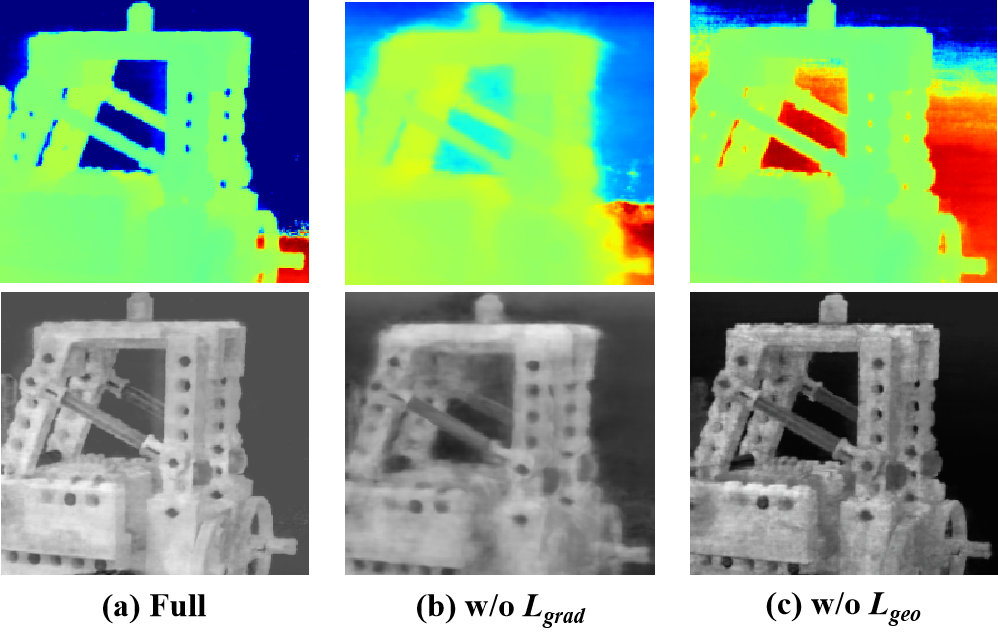}
    \caption{Qualitative Results of ablations on synthetic data.}
    \label{fig:abl1}
    \vspace{-0.3cm}
\end{figure}

\begin{figure}[t]
    \centering
    \includegraphics[width=1\linewidth]{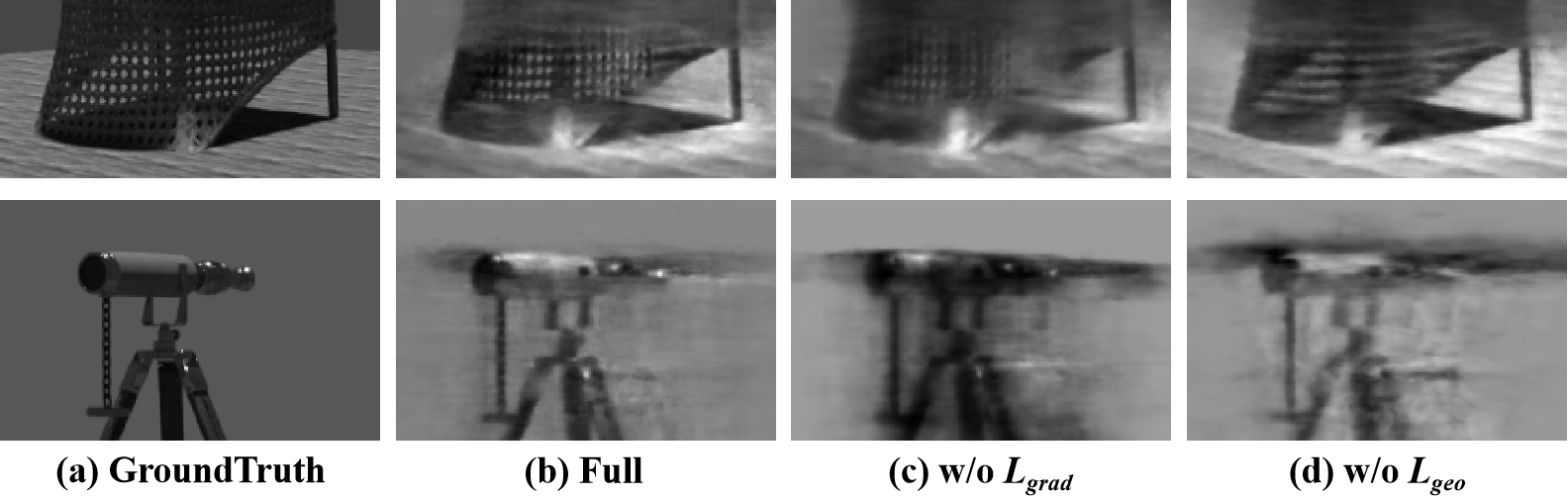}
    \caption{Qualitative Results of ablations on realistic data.}
    \label{fig:abl2}
\end{figure}

\subsection{Ablation studies}
We ablate the work's main contributions, including two prior-based loss functions and the density-based patch sampling strategy on synthetic and realistic event data. The quantitative results of losses on the "Lego" scene are shown in Table~\ref{tab.abl1}. We also indicate the qualitative comparisons in Figure~\ref{fig:abl1}. Clearly, the gradient loss enables the model to correctly learn detailed local structures, including holes or poles. While the geometry loss optimizes global transmittance fields with less noise. Figure~\ref{fig:abl2} gives ablation examples on two realistic event data, namely "wooden chair and "telescope". It can be seen that the full model maintains the best complicated geometries and local structures such as poles and hinged joints.

\begin{figure}
    \centering
    \includegraphics[width=1\linewidth]{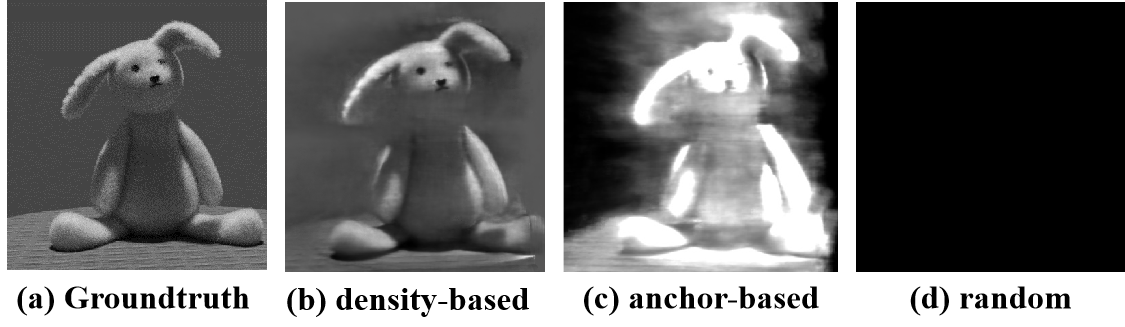}
    \vspace{-0.4cm}
    \caption{Ablation studies of different sampling strategies.}
    \label{fig:sample}
    \vspace{-0.4cm}
\end{figure}

To evaluate the sampling strategy, we did three counterparts containing our density-based patch sampling, random patch sampling, and anchor-based patch sampling. Notably, our gradient loss must require patches, therefore single-pixel sampling strategies such as that in EventNeRF cannot be applied. For a fair comparison, we instead use anchor-based sampling to replace, which means that we first sample several independent pixels by the EventNeRF's sampling strategy. Then we extend these single pixels as patches. Qualitative comparisons are listed in Figure~\ref{fig:sample}. The random patch sampling results in complete failure because of the spatial sparsity of event data. The anchor-based method can resolve the influence of sparsity, however, it only considers the distributions at the centers of patches but ignores the sparsity over whole patches. Our density-based patch sampling outperforms the counterparts and recovers fine-grained local structures. 

\subsection{Efficiency} The proposed method learns from event data to represent 3D scenes in a more efficient way, especially when temporal event distribution is sparse. We evaluate our model and EventNeRF with different maximum window sizes to compare the learning efficiencies on different levels of sparsities. We observe that EventNeRF is not robust to the window size selection. For certain scenarios, it requires a fine hyperparameter search to ensure convergence, while our method can perform relatively well regardless of which window size we choose. Moreover, our approach converges faster than its counterpart. EventNeRF consumes about 500,000 iterations for full convergence, which approximately needs 20 hours, whereas ours only costs 200,000 steps and the time required is about 10 hours. 

\section{Conclusion}
In this work, we present a novel paradigm to reconstruct NeRF from event data with strong physical augmentations by motion, geometry, and density priors. In detail, we bridge motion and geometry priors with NeRF training via the event warping field and the deterministic event generation model. Additionally, we propose the density-guided patch sampling strategy to enable the model to train more efficiently. Furthermore, we propose the first large dataset for event-based 3D representation with high-quality image labels, which contains 101 objects with various materials and geometries. This dataset contributes significantly to advancing research in the related field. We evaluate our approach on both synthetic and real datasets and it is clear that our approach is much superior to the counterpart, especially on realistic and noisy event data. Moreover, the proposed method converges more than 2 times faster. However, this method also has some limitations. For example, the performance depends on the results of the prior extraction branch. In the future, we plan to unify the two stages into a whole optimization formular to optimize the two branches simultaneously. Besides, with the help of large datasets, it is possible to explore generalizable event-based 3D representation without the need for per-scene optimization.

\clearpage
\bibliography{root}

\begin{thebibliography}{10}

\bibitem{achlioptasLearningRepresentationsGenerative2018}
Panos Achlioptas, Olga Diamanti, Ioannis Mitliagkas, and Leonidas Guibas.
\newblock Learning representations and generative models for 3d point clouds.
\newblock In {\em International conference on machine learning}, pages 40--49. PMLR, 2018.

\bibitem{byravan2023nerf2real}
Arunkumar Byravan, Jan Humplik, Leonard Hasenclever, Arthur Brussee, Francesco Nori, Tuomas Haarnoja, Ben Moran, Steven Bohez, Fereshteh Sadeghi, Bojan Vujatovic, et~al.
\newblock Nerf2real: Sim2real transfer of vision-guided bipedal motion skills using neural radiance fields.
\newblock In {\em 2023 IEEE International Conference on Robotics and Automation (ICRA)}, pages 9362--9369. IEEE, 2023.

\bibitem{camunas2014use}
Luis~A Camu{\~n}as-Mesa, Teresa Serrano-Gotarredona, Sio~H Ieng, Ryad~B Benosman, and Bernabe Linares-Barranco.
\newblock On the use of orientation filters for 3d reconstruction in event-driven stereo vision.
\newblock {\em Frontiers in neuroscience}, 8:48, 2014.

\bibitem{chung2023orbeez}
Chi-Ming Chung, Yang-Che Tseng, Ya-Ching Hsu, Xiang-Qian Shi, Yun-Hung Hua, Jia-Fong Yeh, Wen-Chin Chen, Yi-Ting Chen, and Winston~H Hsu.
\newblock Orbeez-slam: A real-time monocular visual slam with orb features and nerf-realized mapping.
\newblock In {\em 2023 IEEE International Conference on Robotics and Automation (ICRA)}, pages 9400--9406. IEEE, 2023.

\bibitem{cui2022dense}
Mingyue Cui, Yuzhang Zhu, Yechang Liu, Yunchao Liu, Gang Chen, and Kai Huang.
\newblock Dense depth-map estimation based on fusion of event camera and sparse lidar.
\newblock {\em IEEE Transactions on Instrumentation and Measurement}, 71:1--11, 2022.

\bibitem{dai2023graspnerf}
Qiyu Dai, Yan Zhu, Yiran Geng, Ciyu Ruan, Jiazhao Zhang, and He~Wang.
\newblock Graspnerf: multiview-based 6-dof grasp detection for transparent and specular objects using generalizable nerf.
\newblock In {\em 2023 IEEE International Conference on Robotics and Automation (ICRA)}, pages 1757--1763. IEEE, 2023.

\bibitem{v2e}
Tobi Delbruck, Yuhuang Hu, and Zhe He.
\newblock V2e: From video frames to realistic dvs event camera streams.
\newblock {\em arXiv e-prints}, pages arXiv--2006, 2020.

\bibitem{gallego2018unifying}
Guillermo Gallego, Henri Rebecq, and Davide Scaramuzza.
\newblock A unifying contrast maximization framework for event cameras, with applications to motion, depth, and optical flow estimation.
\newblock In {\em Proceedings of the IEEE conference on computer vision and pattern recognition}, pages 3867--3876, 2018.

\bibitem{hadviger2021feature}
Antea Hadviger, Igor Cvi{\v{s}}i{\'c}, Ivan Markovi{\'c}, Sacha Vra{\v{z}}i{\'c}, and Ivan Petrovi{\'c}.
\newblock Feature-based event stereo visual odometry.
\newblock In {\em 2021 European Conference on Mobile Robots (ECMR)}, pages 1--6. IEEE, 2021.

\bibitem{hwang2023ev}
Inwoo Hwang, Junho Kim, and Young~Min Kim.
\newblock Ev-nerf: Event based neural radiance field.
\newblock In {\em Proceedings of the IEEE/CVF Winter Conference on Applications of Computer Vision}, pages 837--847, 2023.

\bibitem{kim2016real}
Hanme Kim, Stefan Leutenegger, and Andrew~J Davison.
\newblock Real-time 3d reconstruction and 6-dof tracking with an event camera.
\newblock In {\em Computer Vision--ECCV 2016: 14th European Conference, Amsterdam, The Netherlands, October 11-14, 2016, Proceedings, Part VI 14}, pages 349--364. Springer, 2016.

\bibitem{kogler2011event}
Jurgen Kogler, Martin Humenberger, and Christoph Sulzbachner.
\newblock Event-based stereo matching approaches for frameless address event stereo data.
\newblock In {\em Advances in Visual Computing: 7th International Symposium, ISVC 2011, Las Vegas, NV, USA, September 26-28, 2011. Proceedings, Part I 7}, pages 674--685. Springer, 2011.

\bibitem{kueng2016low}
Beat Kueng, Elias Mueggler, Guillermo Gallego, and Davide Scaramuzza.
\newblock Low-latency visual odometry using event-based feature tracks.
\newblock In {\em 2016 IEEE/RSJ International Conference on Intelligent Robots and Systems (IROS)}, pages 16--23. IEEE, 2016.

\bibitem{lee_uncertainty_2022}
Soomin Lee, Le~Chen, Jiahao Wang, Alexander Liniger, Suryansh Kumar, and Fisher Yu.
\newblock Uncertainty guided policy for active robotic 3d reconstruction using neural radiance fields.
\newblock {\em IEEE Robotics and Automation Letters}, 7(4):12070--12077, 2022.

\bibitem{levoy2023light}
Marc Levoy and Pat Hanrahan.
\newblock Light field rendering.
\newblock In {\em Seminal Graphics Papers: Pushing the Boundaries, Volume 2}, pages 441--452, 2023.

\bibitem{lin2023parallel}
Yunzhi Lin, Thomas M{\"u}ller, Jonathan Tremblay, Bowen Wen, Stephen Tyree, Alex Evans, Patricio~A Vela, and Stan Birchfield.
\newblock Parallel inversion of neural radiance fields for robust pose estimation.
\newblock In {\em 2023 IEEE International Conference on Robotics and Automation (ICRA)}, pages 9377--9384. IEEE, 2023.

\bibitem{liu2023nerf}
Jianlin Liu, Qiang Nie, Yong Liu, and Chengjie Wang.
\newblock Nerf-loc: Visual localization with conditional neural radiance field.
\newblock {\em arXiv preprint arXiv:2304.07979}, 2023.

\bibitem{liuNeuralRenderingReenactment2019}
Lingjie Liu, Weipeng Xu, Michael Zollhoefer, Hyeongwoo Kim, Florian Bernard, Marc Habermann, Wenping Wang, and Christian Theobalt.
\newblock Neural rendering and reenactment of human actor videos.
\newblock {\em ACM Transactions on Graphics (TOG)}, 38(5):1--14, 2019.

\bibitem{liu_dist_2020}
Shaohui Liu, Yinda Zhang, Songyou Peng, Boxin Shi, Marc Pollefeys, and Zhaopeng Cui.
\newblock Dist: Rendering deep implicit signed distance function with differentiable sphere tracing.
\newblock In {\em Proceedings of the IEEE/CVF Conference on Computer Vision and Pattern Recognition}, pages 2019--2028, 2020.

\bibitem{liuGeneralDifferentiableMesh2020}
Shichen Liu, Tianye Li, Weikai Chen, and Hao Li.
\newblock A general differentiable mesh renderer for image-based {{3D}} reasoning.
\newblock {\em IEEE Transactions on Pattern Analysis and Machine Intelligence}, 44(1):50--62, 2020.

\bibitem{liu_learning_2019}
Shichen Liu, Shunsuke Saito, Weikai Chen, and Hao Li.
\newblock Learning to infer implicit surfaces without 3d supervision.
\newblock {\em Advances in Neural Information Processing Systems}, 32, 2019.

\bibitem{lombardiNeuralVolumesLearning2019}
Stephen Lombardi, Tomas Simon, Jason Saragih, Gabriel Schwartz, Andreas Lehrmann, and Yaser Sheikh.
\newblock Neural volumes: learning dynamic renderable volumes from images.
\newblock {\em ACM Transactions on Graphics (TOG)}, 38(4):1--14, 2019.

\bibitem{lombardi_neural_2019}
Stephen Lombardi, Tomas Simon, Jason Saragih, Gabriel Schwartz, Andreas Lehrmann, and Yaser Sheikh.
\newblock Neural volumes: Learning dynamic renderable volumes from images.
\newblock {\em arXiv preprint arXiv:1906.07751}, 2019.

\bibitem{maggio2023loc}
Dominic Maggio, Marcus Abate, Jingnan Shi, Courtney Mario, and Luca Carlone.
\newblock Loc-nerf: Monte carlo localization using neural radiance fields.
\newblock In {\em 2023 IEEE International Conference on Robotics and Automation (ICRA)}, pages 4018--4025. IEEE, 2023.

\bibitem{mildenhall_nerf_2022}
Ben Mildenhall, Pratul~P Srinivasan, Matthew Tancik, Jonathan~T Barron, Ravi Ramamoorthi, and Ren Ng.
\newblock Nerf: Representing scenes as neural radiance fields for view synthesis.
\newblock {\em Communications of the ACM}, 65(1):99--106, 2021.

\bibitem{nehvi2021differentiable}
Jalees Nehvi, Vladislav Golyanik, Franziska Mueller, Hans-Peter Seidel, Mohamed Elgharib, and Christian Theobalt.
\newblock Differentiable event stream simulator for non-rigid 3d tracking.
\newblock In {\em Proceedings of the IEEE/CVF Conference on Computer Vision and Pattern Recognition}, pages 1302--1311, 2021.

\bibitem{niemeyer_differentiable_2020}
Michael Niemeyer, Lars Mescheder, Michael Oechsle, and Andreas Geiger.
\newblock Differentiable volumetric rendering: Learning implicit 3d representations without 3d supervision.
\newblock In {\em Proceedings of the IEEE/CVF Conference on Computer Vision and Pattern Recognition}, pages 3504--3515, 2020.

\bibitem{pan2019bringing}
Liyuan Pan, Cedric Scheerlinck, Xin Yu, Richard Hartley, Miaomiao Liu, and Yuchao Dai.
\newblock Bringing a blurry frame alive at high frame-rate with an event camera.
\newblock In {\em Proceedings of the IEEE/CVF Conference on Computer Vision and Pattern Recognition}, pages 6820--6829, 2019.

\bibitem{qiPointnetDeepLearning2017}
Charles~R. Qi, Hao Su, Kaichun Mo, and Leonidas~J. Guibas.
\newblock Pointnet: {{Deep}} learning on point sets for 3d classification and segmentation.
\newblock In {\em Proceedings of the {{IEEE}} Conference on Computer Vision and Pattern Recognition}, pages 652--660, 2017.

\bibitem{rebecq2018emvs}
Henri Rebecq, Guillermo Gallego, Elias Mueggler, and Davide Scaramuzza.
\newblock Emvs: Event-based multi-view stereo—3d reconstruction with an event camera in real-time.
\newblock {\em International Journal of Computer Vision}, 126(12):1394--1414, 2018.

\bibitem{rebecq2016evo}
Henri Rebecq, Timo Horstsch{\"a}fer, Guillermo Gallego, and Davide Scaramuzza.
\newblock Evo: A geometric approach to event-based 6-dof parallel tracking and mapping in real time.
\newblock {\em IEEE Robotics and Automation Letters}, 2(2):593--600, 2016.

\bibitem{rebecq2019high}
Henri Rebecq, Ren{\'e} Ranftl, Vladlen Koltun, and Davide Scaramuzza.
\newblock High speed and high dynamic range video with an event camera.
\newblock {\em IEEE transactions on pattern analysis and machine intelligence}, 43(6):1964--1980, 2019.

\bibitem{rogister2011asynchronous}
Paul Rogister, Ryad Benosman, Sio-Hoi Ieng, Patrick Lichtsteiner, and Tobi Delbruck.
\newblock Asynchronous event-based binocular stereo matching.
\newblock {\em IEEE Transactions on Neural Networks and Learning Systems}, 23(2):347--353, 2011.

\bibitem{romero2022embodied}
Javier Romero, Dimitrios Tzionas, and Michael~J Black.
\newblock Embodied hands: Modeling and capturing hands and bodies together.
\newblock {\em arXiv preprint arXiv:2201.02610}, 2022.

\bibitem{rudnev_eventnerf_nodate}
Viktor Rudnev, Mohamed Elgharib, Christian Theobalt, and Vladislav Golyanik.
\newblock Eventnerf: Neural radiance fields from a single color event camera.
\newblock In {\em Proceedings of the IEEE/CVF Conference on Computer Vision and Pattern Recognition}, pages 4992--5002, 2023.

\bibitem{sitzmannDeepvoxelsLearningPersistent2019}
Vincent Sitzmann, Justus Thies, Felix Heide, Matthias Nießner, Gordon Wetzstein, and Michael Zollhofer.
\newblock Deepvoxels: {{Learning}} persistent 3d feature embeddings.
\newblock In {\em Proceedings of the {{IEEE}}/{{CVF Conference}} on {{Computer Vision}} and {{Pattern Recognition}}}, pages 2437--2446, 2019.

\bibitem{sunderhauf2023density}
Niko S{\"u}nderhauf, Jad Abou-Chakra, and Dimity Miller.
\newblock Density-aware nerf ensembles: Quantifying predictive uncertainty in neural radiance fields.
\newblock In {\em 2023 IEEE International Conference on Robotics and Automation (ICRA)}, pages 9370--9376. IEEE, 2023.

\bibitem{tancik_block-nerf_2022}
Matthew Tancik, Vincent Casser, Xinchen Yan, Sabeek Pradhan, Ben Mildenhall, Pratul~P Srinivasan, Jonathan~T Barron, and Henrik Kretzschmar.
\newblock Block-nerf: Scalable large scene neural view synthesis.
\newblock In {\em Proceedings of the IEEE/CVF Conference on Computer Vision and Pattern Recognition}, pages 8248--8258, 2022.

\bibitem{thies_deferred_2019}
Justus Thies, Michael Zollh{\"o}fer, and Matthias Nie{\ss}ner.
\newblock Deferred neural rendering: Image synthesis using neural textures.
\newblock {\em Acm Transactions on Graphics (TOG)}, 38(4):1--12, 2019.

\bibitem{thiesDeferredNeuralRendering2019}
Justus Thies, Michael Zollhöfer, and Matthias Nießner.
\newblock Deferred neural rendering: {{Image}} synthesis using neural textures.
\newblock {\em Acm Transactions on Graphics (TOG)}, 38(4):1--12, 2019.

\bibitem{tong2023enforcing}
Mukun Tong, Charles Dawson, and Chuchu Fan.
\newblock Enforcing safety for vision-based controllers via control barrier functions and neural radiance fields.
\newblock In {\em 2023 IEEE International Conference on Robotics and Automation (ICRA)}, pages 10511--10517. IEEE, 2023.

\bibitem{Tosi_2023_CVPR}
Fabio Tosi, Alessio Tonioni, Daniele De~Gregorio, and Matteo Poggi.
\newblock Nerf-supervised deep stereo.
\newblock In {\em Conference on Computer Vision and Pattern Recognition (CVPR)}, pages 855--866, June 2023.

\bibitem{vidal2018ultimate}
Antoni~Rosinol Vidal, Henri Rebecq, Timo Horstschaefer, and Davide Scaramuzza.
\newblock Ultimate slam? combining events, images, and imu for robust visual slam in hdr and high-speed scenarios.
\newblock {\em IEEE Robotics and Automation Letters}, 3(2):994--1001, 2018.

\bibitem{wang_co-slam_nodate}
Hengyi Wang, Jingwen Wang, and Lourdes Agapito.
\newblock Co-slam: Joint coordinate and sparse parametric encodings for neural real-time slam.
\newblock In {\em Proceedings of the IEEE/CVF Conference on Computer Vision and Pattern Recognition}, pages 13293--13302, 2023.

\bibitem{wangPixel2meshGenerating3d2018}
Nanyang Wang, Yinda Zhang, Zhuwen Li, Yanwei Fu, Wei Liu, and Yu-Gang Jiang.
\newblock Pixel2mesh: {{Generating}} 3d mesh models from single rgb images.
\newblock In {\em Proceedings of the {{European}} Conference on Computer Vision ({{ECCV}})}, pages 52--67, 2018.

\bibitem{wang2022evac3d}
Ziyun Wang, Kenneth Chaney, and Kostas Daniilidis.
\newblock Evac3d: From event-based apparent contours to 3d models via continuous visual hulls.
\newblock In {\em European conference on computer vision}, pages 284--299. Springer, 2022.

\bibitem{xiangli_bungeenerf_2023}
Yuanbo Xiangli, Linning Xu, Xingang Pan, Nanxuan Zhao, Anyi Rao, Christian Theobalt, Bo~Dai, and Dahua Lin.
\newblock Bungeenerf: Progressive neural radiance field for extreme multi-scale scene rendering.
\newblock In {\em European conference on computer vision}, pages 106--122. Springer, 2022.

\bibitem{xu2020eventcap}
Lan Xu, Weipeng Xu, Vladislav Golyanik, Marc Habermann, Lu~Fang, and Christian Theobalt.
\newblock Eventcap: Monocular 3d capture of high-speed human motions using an event camera.
\newblock In {\em Proceedings of the IEEE/CVF Conference on Computer Vision and Pattern Recognition}, pages 4968--4978, 2020.

\bibitem{mi_switch-nerf_2023}
MI~Zhenxing and Dan Xu.
\newblock Switch-nerf: Learning scene decomposition with mixture of experts for large-scale neural radiance fields.
\newblock In {\em The Eleventh International Conference on Learning Representations}, 2022.

\bibitem{zhou_nerf_nodate}
Allan Zhou, Moo~Jin Kim, Lirui Wang, Pete Florence, and Chelsea Finn.
\newblock Nerf in the palm of your hand: Corrective augmentation for robotics via novel-view synthesis.
\newblock In {\em Proceedings of the IEEE/CVF Conference on Computer Vision and Pattern Recognition}, pages 17907--17917, 2023.

\bibitem{zhou_stereo_2018}
Tinghui Zhou, Richard Tucker, John Flynn, Graham Fyffe, and Noah Snavely.
\newblock Stereo magnification: Learning view synthesis using multiplane images.
\newblock {\em arXiv preprint arXiv:1805.09817}, 2018.

\bibitem{zhou2018semi}
Yi~Zhou, Guillermo Gallego, Henri Rebecq, Laurent Kneip, Hongdong Li, and Davide Scaramuzza.
\newblock Semi-dense 3d reconstruction with a stereo event camera.
\newblock In {\em Proceedings of the European conference on computer vision (ECCV)}, pages 235--251, 2018.

\bibitem{zhou2021event}
Yi~Zhou, Guillermo Gallego, and Shaojie Shen.
\newblock Event-based stereo visual odometry.
\newblock {\em IEEE Transactions on Robotics}, 37(5):1433--1450, 2021.

\bibitem{zhu2023latitude}
Zhenxin Zhu, Yuantao Chen, Zirui Wu, Chao Hou, Yongliang Shi, Chuxuan Li, Pengfei Li, Hao Zhao, and Guyue Zhou.
\newblock Latitude: Robotic global localization with truncated dynamic low-pass filter in city-scale nerf.
\newblock In {\em 2023 IEEE International Conference on Robotics and Automation (ICRA)}, pages 8326--8332. IEEE, 2023.

\bibitem{zhu_nicer-slam_2023}
Zihan Zhu, Songyou Peng, Viktor Larsson, Zhaopeng Cui, Martin~R Oswald, Andreas Geiger, and Marc Pollefeys.
\newblock Nicer-slam: Neural implicit scene encoding for rgb slam.
\newblock {\em arXiv preprint arXiv:2302.03594}, 2023.

\bibitem{zuo2022devo}
Yi-Fan Zuo, Jiaqi Yang, Jiaben Chen, Xia Wang, Yifu Wang, and Laurent Kneip.
\newblock Devo: Depth-event camera visual odometry in challenging conditions.
\newblock In {\em 2022 International Conference on Robotics and Automation (ICRA)}, pages 2179--2185. IEEE, 2022.

\end{thebibliography}

\end{document}